# Eff-3DPSeg: 3D organ-level plant shoot segmentation using annotation-efficient point clouds

Liyi Luo,[1] Xintong Jiang,[1] Yu Yang,[1,2] Eugene Roy Antony Samy,[1] Mark Lefsrud,[1] Valerio Hoyos-Villegas[3], and Shangpeng Sun[1]*

[1] Bioresource Engineering Department, McGill University, Montreal, QC, Canada

[2] Key Laboratory of Advanced Process Control for Light Industry (Ministry of Education), Jiangnan University, China

[3] Plant Science Department, McGill University, Montreal, QC, Canada,

*Corresponding author. Email: shangpeng.sun@mcgill.ca

**Abstract**

Reliable and automated 3D plant shoot segmentation is a core prerequisite for the extraction of plant phenotypic traits at the organ level. Combining deep learning and point clouds can provide effective ways to address the challenge. However, fully supervised deep learning methods require datasets to be point-wise annotated, which is extremely expensive and time-consuming. In our work, we proposed a novel weakly supervised framework, Eff-3DPSeg, for 3D plant shoot segmentation. First, high-resolution point clouds of soybean were reconstructed using a low-cost photogrammetry system, and the Meshlab-based Plant Annotator was developed for plant point cloud annotation. Second, a weakly-supervised deep learning method was proposed for plant organ segmentation. The method contained: (1) Pretraining a self-supervised network using Viewpoint Bottleneck loss to learn meaningful intrinsic structure representation from the raw point clouds; (2) Fine-tuning the pre-trained model with about only 0.5% points being annotated to implement plant organ segmentation. After, three phenotypic traits (stem diameter, leaf width, and leaf length) were extracted. To test the generality of the proposed method, the public dataset Pheno4D was included in this study. Experimental results showed that the weakly-supervised network obtained similar segmentation performance compared with the fully-supervised setting. Our method achieved 95.1%, 96.6%, 95.8% and 92.2% in the Precision, Recall, F1-score, and mIoU for stem leaf segmentation and 53%, 62.8% and 70.3% in the AP, AP@25, and AP@50 for leaf instance segmentation. This study provides an effective way for characterizing 3D plant architecture, which will become useful for plant breeders to enhance selection processes.

## 1. Introduction

High-throughput plant phenotyping is crucial to improve the understanding of the interactions between plant genotypes and phenotypes, which can be highly useful to speed up the selection of desired genotypes [1]. Traditional manual methods for plant phenotyping are highly labor-intensive, time-consuming, and prone to be inaccurate [2]. High throughput plant phenotyping technologies have been identified as a bottleneck limiting systematic studies of plant gene functions and plant multi-omics research [3]. Recently, computer vision-based methods are gaining increased attention among plant researchers for efficiently visualizing plant architecture, measuring phenotypic traits, and reducing human errors.

Two-dimensional (2D) image-based methods have been used widely for high-throughput plant phenotyping during the last several decades [4]. For example, existing studies have demonstrated extracting multiple phenotypic traits from 2D images for a wide range of crops, such as tomato, maize, sorghum, wheat and so on [5-7]. However, all these 2D imaging technologies have significant drawbacks: (1) it is difficult to address the occlusion issues due to the lack of depth information, and (2) it is difficult to determine object structure information [8].

To address the disadvantages of 2D image-based methods, much effort has been made in the development of 3D imaging systems for plant phenotyping in the past decade [9]. Compared with 2D data, 3D data not only greatly address the aforementioned limitations, but also provide opportunities to extract new and more complex phenotypic traits by generating accurate coordinates and distance estimates of objects [10]. At present, rapid 3D plant data acquisition benefits from the development of sensing technology and improvement of computational performance [11, 12]. For 3D plant data acquisition, light detection and ranging (LiDAR) [11, 13], Time-of-Flight (ToF) [14], depth cameras [15], and Multi-view Stereo (MVS) cameras [16, 17] are widely used for plant 3D reconstruction and phenotypic analysis.

After acquiring the precise plant point clouds, reliable and automated plant organ segmentation becomes a premise for phenotypic analysis. Many existing studies developed traditional computer vision methods for plant organ segmentation from 3D data, such as threshold-based methods [18], geometry-based methods [19], the octree-based methods [20], and 3D skeleton-based methods [21]. These methods can handle several types of plants with simple structures through tedious and labor-intensive parameter tuning, which cannot be suitable for big-data processing requirements in high-throughput plant phenotyping [22]. Recently, there has been rapid growth in the field of deep learning-based methods which can improve the generality and accuracy of plant phenotypic analysis. For example, Shi et al. [23] applied a deep learning method to segment point clouds of whole plants into organs. Li et al. [24] developed a network named DeepSeg3DMaize for plant point cloud segmentation that integrated high-throughput data acquisition and deep learning. PlantNet [25], a dual-function 3D deep learning network, simultaneously implemented stem-leaf semantic and instance segmentation for three different species of crops. However, training such deep learning models is still challenging. Firstly, all these fully-supervised deep learning methods need datasets with point-wise annotation for model training. Point-wise annotation of large-scale 3D plant point clouds is very time-consuming, and a user-friendly toolkit for annotation is currently absent. To this end, one of the potential solutions is to develop deep learning architectures with learning models using a tiny subset for annotations. This is usually referred to as weakly-supervised representation learning, which learns meaningful representations from data without the usage of annotation through the self-supervised learning method, and then fine-tunes the pre-trained network with weakly-supervised datasets. This learning strategy could produce a better performance than the training models using a weak supervision directly. For example, PointContrast [26] proposed a PointInfoNCE loss and verified its effectiveness on a set of weakly-supervised 3D scene understanding tasks. Contrastive Scene Context (CSC) [27] introduced a loss function that contrasted features aggregated in local partitions. Secondly, there are few large-scale well-labeled plant point cloud datasets and there is no universal benchmark data for plant organ instance segmentation. Pheno4D [28] is a spatio-temporal point cloud dataset, but only includes seven tomato and seven maize plants. ROSE-X [29] provided an annotated 3D dataset of rosebush plants for training and evaluation of organ segmentation methods. However, this dataset only contained 11 annotated 3D plant models with organ labels for voxels corresponding to the plant shoot. Using numerous and high-quality raw 3D plant dataset is beneficial for deep learning models training to obtain better results. Thus, building a large-scale well-labeled 3D plant point cloud dataset is the key to the deep learning-based high-throughput plant phenotyping. In this work, we proposed a weakly-supervised deep learning-based framework, Eff-3DPSeg, for both plant stem-leaf segmentation and leaf instance segmentation. To do so, a low-cost Multi-view Stereo Pheno

Platform (MVSP2) was developed to acquire point clouds for individual plants, and then a point cloud annotation tool, Meshlab-based Plant Annotator (MPA), was used for the data annotation. After that, a weakly-supervised deep learning network was developed for an end-to-end 3D plant architecture segmentation. Finally, three plant phenotypic traits were extracted based on the segmentation results. To summarize, the main contributions are as follows:

1. We proposed a weakly-supervised 3D plant shoot segmentation framework: Eff-3DPSeg, which only uses about 0.5% points being labeled for training the networks. It is a robust deep learning-based method that can be reapplied to other species of plants with minor modifications.

2. We built a large-scale well-labeled soybean spatio-temporal dataset, which includes point clouds in different grow stages among three weeks and organ-level annotations.

3. We demonstrated the effectiveness of the weakly supervised plant organ segmentation methods by comparing the segmentation performance with the fully-supervised method using two types plants: soybean and tomato.

## 2. Materials and Methods

### 2.1 Overview

Overall, the proposed Eff-3DPSeg framework consists of three parts (Fig.1). The first part (A) is for the high-resolution plant point clouds acquisition and annotation. The plant point clouds were reconstructed using a low-cost photogrammetry platform MVSP2, and the point-wise annotation of point clouds was labeled by the tool MPA. The second part (B) is to apply the proposed weakly-supervised deep learning networks for 3D plant shoot segmentation, including plant stem leaf segmentation and leaf instance segmentation. The third part (C) is the plant phenotypic traits extraction. Utilizing the results of plant organ segmentation, we extracted three plant phenotypic traits, including stem diameter, leaf width, and leaf length.

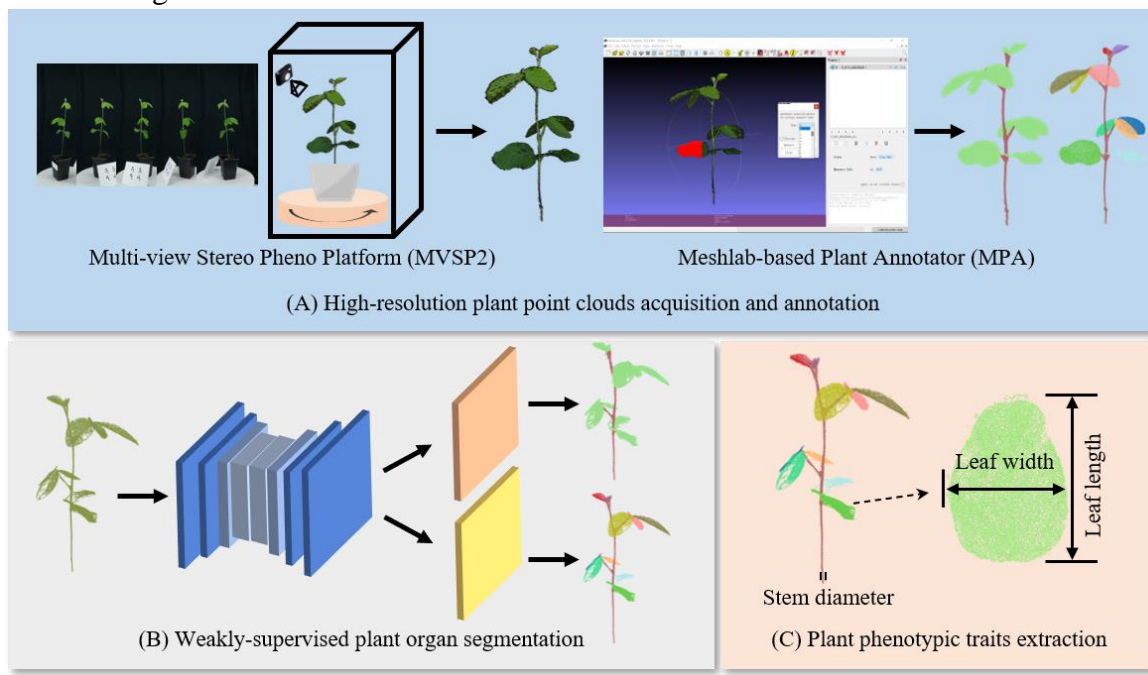

Figure 1: The overall workflow of the proposed Eff-3DSeg framework. (A) the plant point clouds were reconstructed using a Multi-view Stereo Pheno Platform, and the annotation of point clouds were labeled by the Meshlab-based Plant Annotator. (B) the proposed weakly-supervised plant organ segmentation networks for plant stem leaf segmentation and leaf instance segmentation. (C) Three plant phenotypic

Page 3 of 21

traits (stem diameter, leaf width, and leaf length) were extracted using the results of plant organ segmentation.

## 2.2    High-resolution point cloud dataset

### 2.2.1   Soybean point cloud acquisition

The platform MVSP2 mainly consists of an RGB camera (LUMIX DMC-G7W, Panasonic, Japan), a turntable, and LED lights. Soybean seeds were planted in pots which were put in a growth chamber until the seeds were geminated and the unifoliate leaves unfold. Then, plants were moved from the growth chamber to an indoor environment with room temperature (23˚C), and the plants were lighted 16 hours per day with the light intensity of 180~200 µmol/m$^2$/sec using growth lights. For the image data collection, a plant was placed on the turntable with a speed of 360˚/minute, and 60 images were captured for each plant. After that, the software Agisoft Metashape (Agisoft LLC, St. Petersburg, Russia) was used for point cloud reconstruction from the collected images. The imaging data collection was conducted every Monday, Wednesday and Friday from May 6 to May 27, 2022. A total of 30 soybean plants were scanned at the beginning, but two of them died during the image collection period, producing 258 point clouds, Fig. 2 presented representative point clouds of a soybean plant from May 6 to May 27, 2022.

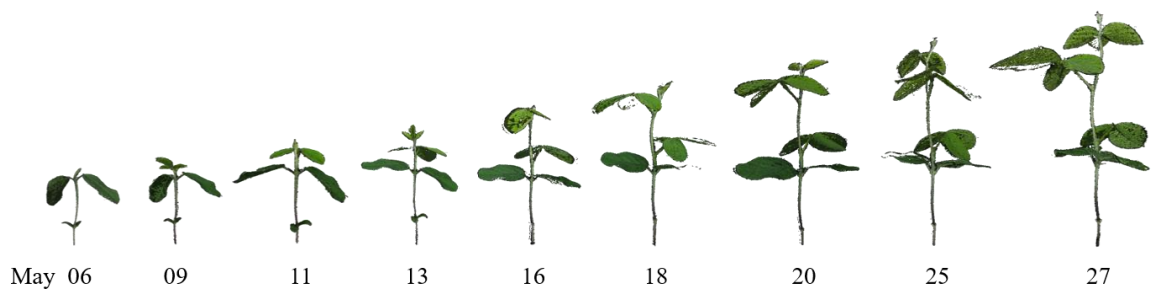

May  06         09           11           13           16           18           20           25           27

Figure 2: Point clouds of a soybean plant captured from May 6 to 27, 2022.

### 2.2.2   Soybean point cloud annotation

We developed a point cloud annotation tool MPA for point cloud annotation. In the processing, we selected the points of an individual organ using the built-in function 'select vertex cluster' in MeshLab; we then assigned a predefined label to the selected points which is indicated using a unique color. After annotation, the original point cloud and labels can be exported from MeshLab. We predefined a total of 70 categories in the tool, which is enough for the organ instance labeling for a soybean plant under different growing stages in this study. We labeled each point as 'stem', or 'leaf', and each leaf had its unique ID label, differentiating it from other leaves in the same plant point cloud (Fig. 3). In this study, we totally annotated 145 plant samples.

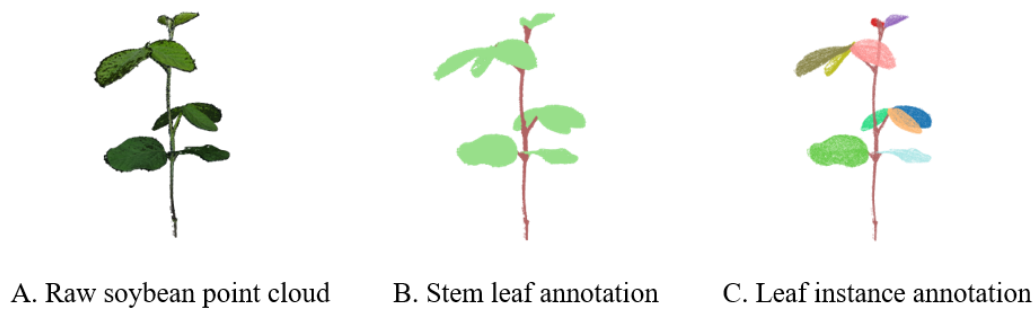

A. Raw soybean point cloud        B. Stem leaf annotation        C. Leaf instance annotation

Figure 3: Representative (A) a raw soybean point cloud and annotation results. (B) stem leaf annotation and (C) leaf instance annotation.

### 2.2.3 Pheno4D dataset

To test the segmentation performance of the proposed 3D deep learning networks, we included another public plant point cloud dataset Pheno4D [28] which had tomato and maize plants. In this study, tomato point clouds in Pheno4D were used because they have more complex structures compared with maize plants. The dataset contained seven tomato plants scanned on 20 different days, generating 140 point clouds with 77 point clouds being annotated. The points were labeled into three categories: 'soil', 'stem', and 'leaf', where each leaf was annotated with a unique label making it distinctive from the other leaves on the same plant. In Table 1, we showed a comparison between our dataset and the Pheno4D.

Table 1: Details of our dataset and the Pheno4D dataset

| Dataset | Total Point clouds | Labeled Point clouds | Extension Period | Measure Frequency | Sensor | Labels |
|---|---|---|---|---|---|---|
| Pheno4D | 224 | 126 | 2-3 weeks | every 2 days | Laser scanner | plant organ instances |
| Soybean (ours) | 258 | 145 | 3 weeks | every 2 or 3 days | RGB camera | plant organ instances |

### 2.2.4 Generation of annotation-efficient dataset

In this study, we explored 3D plant organ segmentation with a limited budget for plant point cloud annotation. We tried three different labeling settings, i.e., annotating 50, 100, and 200 points of each point cloud for the network training. To generate the annotation-efficient dataset, firstly, each plant point cloud was down-sampled with a ratio factor of 0.2. Then, we randomly chose [50, 100, 200] points in each down-sampled point cloud and kept the original labels and set other points' labels to '*None*'. In order to reduce the amount of the calculation and focus on plant organ, we deleted the 'soil' points in Pheno4D.

### 2.3 3D weakly-supervised plant organ segmentation network

### 2.3.1 Overview of the proposed method

There are two main steps for the proposed weakly-supervised plant organ segmentation method (Fig. 4). First, a backbone network is pretrained using a 3D self-supervised representation learning method Viewpoint Bottleneck (VIB) [30, 31]. Then, the pertained model was modified by adding a semantic segmentation head and an instance segmentation head respectively and fine-tuned using the weakly-annotated point clouds to implement the plant stem leaf segmentation and leaf instance segmentation.

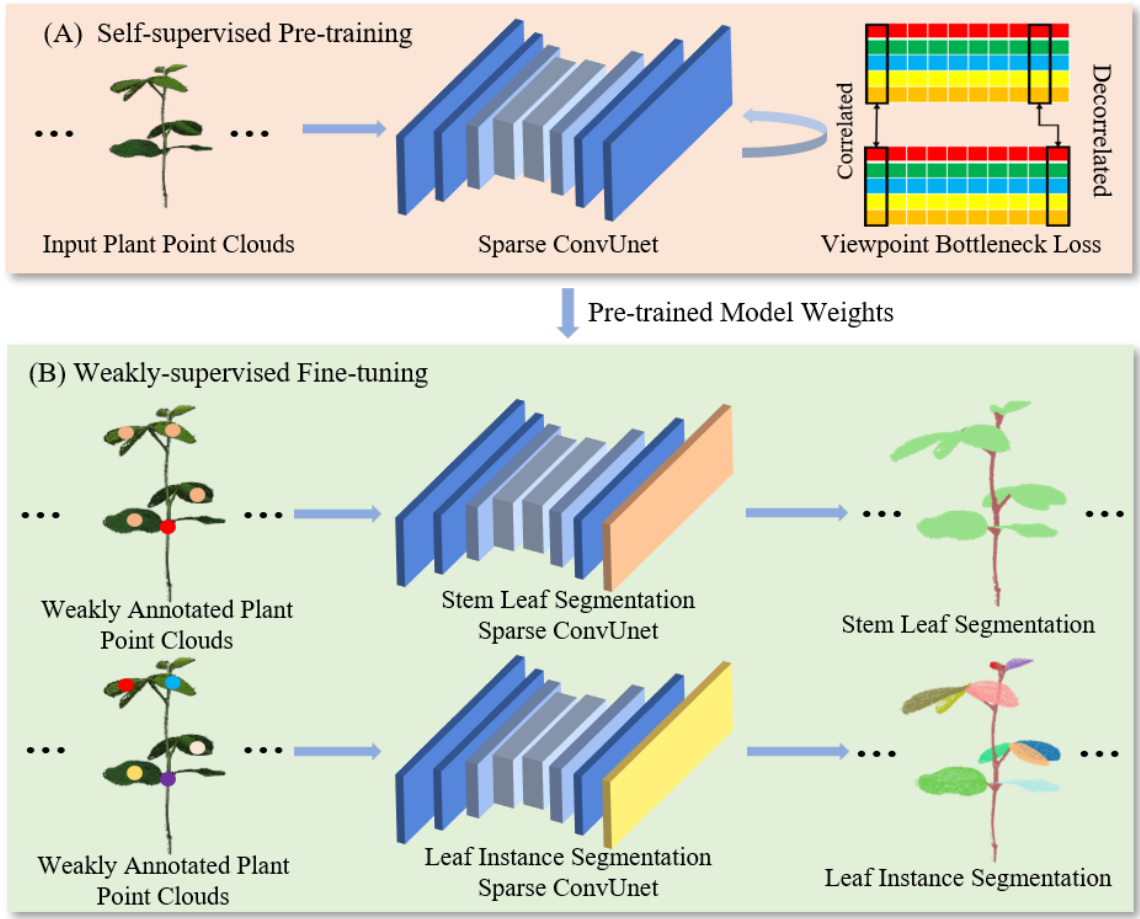

Figure 4: The pipeline of the proposed weakly-supervised plant organ segmentation framework. (A) we firstly pretrained a backbone network with a self-supervised representation learning method using Viewpoint Bottleneck loss function. (B) We modified and fine-tuned the pretrained model for plant stem-leaf segmentation and leaf instance segmentation using the weakly-annotated point clouds.

### 2.3.2 Self-supervised pre-training

The key to effective weakly-supervised learning is leveraging numerous unlabeled points in the plant point clouds. To do so, we applied a self-supervised representation learning method, Viewpoint Bottleneck (VIB), to learn meaningful representations from the plant point clouds without relying on any annotations. As shown in Fig. 5, two viewpoints $X_p(M×6)$ and $X_q(M×6)$ were obtained from an input plant point cloud $X(M×6)$ transformed with random geometric transformations. $M$ is the number of points for the input point cloud, and each point includes $x$, $y$, $z$ coordinates and Red-Green-Blue color intensities. Then two point clouds generated by the two viewpoints were fed into the same Sparse ConvUnet network $f_\theta$ to obtain the representation of the point cloud features [26] (Fig. 6). The outputs from $f_\theta$ were two high-dimensional representations $Z_p(M×D)$ and $Z_q(M× D)$. $D$ is the dimension of the representation of the point cloud. To keep computation tractable, we sampled the representations by Farthest Point Sampling (FPS) from $M×D$ to $H×D$, where $H$ is the number of points after down sampling. Finally, the cross-correlation matrix $Z(D×D)$ was computed on the batch dimension from the two down-sampled representations. The Sparse ConvUnet was used as the backbone to extract point features. This backbone provides discriminative point-wise features for subsequent processing. In addition, it has relatively small GPU memory footprints, which well suits the high-resolution plant 3D data that have large amounts of points to be processed simultaneously and build deeper network to learn more representations of plant point clouds, such as contextual and geometric information.

It was implemented using the MinkowskiEngine [32] which is an open-source auto-differentiation library for sparse tensors to implement the generalized sparse convolution. The network was trained with the VIB loss (Eq. 1):

$$L_{VIB} \triangleq \sum_i (1 - Z_{ii})^2 + \lambda \sum_i \sum_{j \neq i} Z_{ij}^2 \tag{1}$$

where $\lambda$ is a positive constant trading off the two terms of the loss function. VIB loss aims to push diagonal elements $Z_{ii}$ to 1, and off-diagonal elements $Z_{ij}$ to 0. In this way, it maximizes the correlation between corresponding feature channels meanwhile decorrelates different feature channels. As shown in Fig. 5-right, five vectors of different colors demonstrate the sampled representations of point clouds. VIB operates on the feature dimension, to correlate the corresponding channels and decorrelate the different channels.

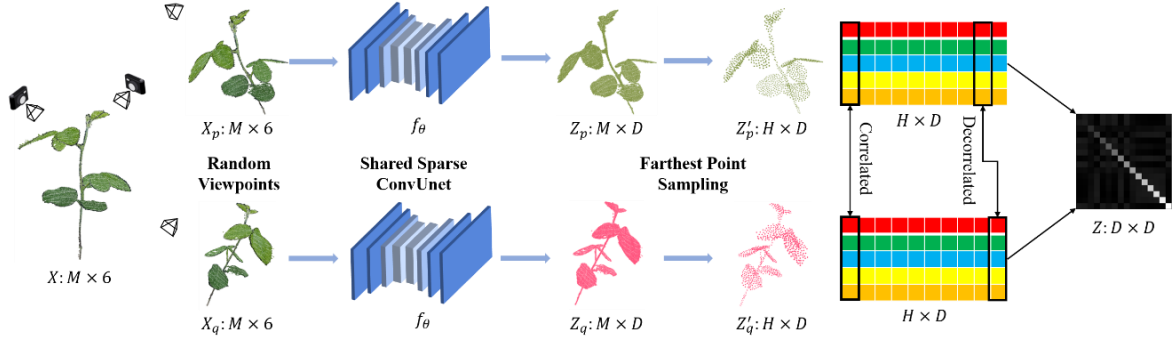

Figure 5: Illustration of the self-supervised pre-training method: Viewpoint Bottleneck. $X$ is a plant point cloud represented by the concatenation of 3D coordinates and colors ($M \times 6$, $M$ is the number of points). After two random geometric transformations, we obtained its two augmentations $X_p$ and $X_q$. They were fed to the shared Sparse ConvUnet $f_\theta$ to obtain two high dimensional representation sets $Z_p$ and $Z_q$ ($M \times D$, $D$ is the number of representation dimension). To keep computation tractable, we applied the Farthest Point Sampling on the representations to get down-sampled representation $Z_P{'}$ and $Z_q{'}$ ($H \times D$), H is the point number of the down-sampled representation. Finally, the viewpoint bottleneck was imposed on the cross-correlation matrix between $Z_P{'}$ and $Z_q{'}$, which was denoted as $Z$.

Overall, through the self-supervised pretraining (Fig. 4(A)), a pre-trained model was learned, which contained meaningful representations leveraged the intrinsic structure between enormous unlabeled points in the plant point clouds.

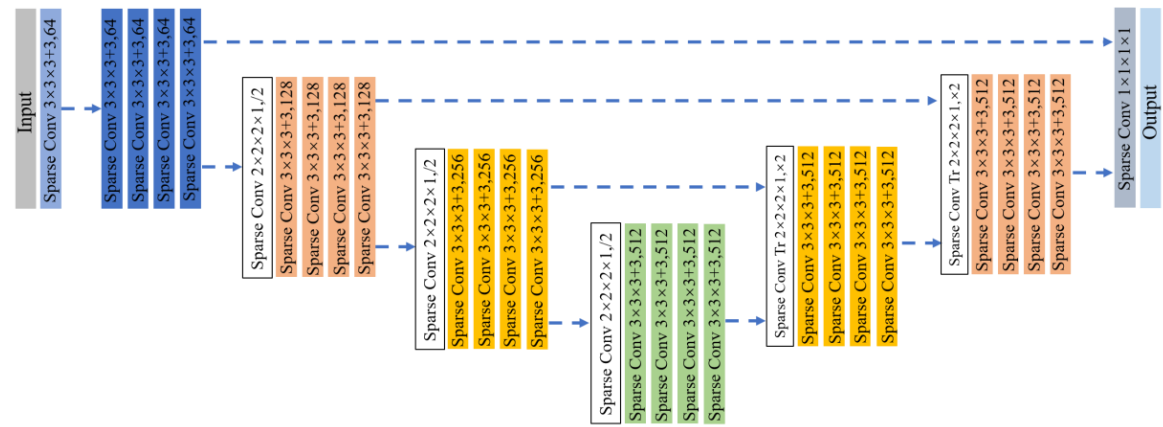

Figure 6: Architecture of Sparse ConvUnet

### 2.3.3 Weakly-supervised plant organ segmentation

After pre-training, we fine-tuned the pre-trained model by adding a stem leaf segmentation head (orange layer in Fig. 4(B)) and a leaf instance segmentation head (yellow layer in Fig. 4(B)) to implement stem leaf segmentation and leaf instance segmentation with the weak-annotation point clouds, respectively. In our experiments, we had three (50, 100, and 200-point) weakly-supervised training settings. In Fig. 4(B), the sparse different color points represented the weak annotations of the plant point cloud. This training scheme is much better than directly training with annotation efficient point clouds, because the feature representation and intrinsic structure between enormous unlabeled points are fully leveraged by self-supervised pre-training. This meaningful representation information could be beneficial for the weakly supervised learning. In our implementation (Fig.7), the backbone network provides discriminative point-wise features $F$ for the subsequent processing.

**Plant stem leaf segmentation:** We utilized a Multi-layer Perceptron (MLP) to produce stem leaf semantic score ($M \times n$) by the point features $F$ for the $M$ points over the $n$ classes. And then the predicted stem leaf labels for each point were obtained through Argmax operation. The plant stem leaf segmentation Sparse ConvUnet was optimized by a cross-entropy loss.

**Plant leaf instance segmentation:** The leaf instance segmentation refers to the task of not only assign to every point a semantic label, but also an instance ID of each leaf. In our implementation, we fed the point features $F$ into two branches. One of the branches is the same as stem leaf segmentation, achieving stem leaf semantic labels to select 'leaf' points for individual leaf clustering. The other branch is called offset branch for predicting a point-wise offset vector $O$ to shift original coordinate $Coords$ towards the $Shift\ Coords$. The offset module was implemented by two sparse convolutional layers and a Batch Normalization layer. We used a clustering method [33] to group points into candidate clusters on stem leaf semantic labels $S$ and dual coordinate sets, original $Coords$ and $Shift\ Coords$, which produced $Cc$ and $Cs$, respectively. In the clustering method, we used 1.5mm-ball as the threshold for every point to find its neighboring points. The threshold was selected based on the distances between points of soybean and tomato point clouds. Within the ball, the points are grouped into one individual leaf when they have the 'leaf' label. The choice of the radius of the ball is affected by the point density of point clouds. Lastly, we obtained the final clustering results $C$ as the union of $Cc$ and $Cs$ [32]. We trained the whole leaf-instance segmentation network with the voting-center loss including three parts [27,32]:

$$\text{Loss}_{is} = L_{sem} + L_{o-reg} + L_{o-dir} \tag{3}$$

where $L_{sem}$ is a cross-entropy loss, $L_{o\text{-}reg}$, and $L_{o\text{-}dir}$ are losses of the offset prediction:

$$L_{o\_reg} = \frac{1}{\sum_i m_i} \sum_i \| o_i - (\hat{c}_i - p_i) \| \cdot m_i \tag{4}$$

$$L_{o\_dir} = -\frac{1}{\sum_i m_i} \sum_i \frac{o_i}{\|o_i\|_2} \cdot \frac{\hat{c}_i - p_i}{\|\hat{c}_i - p_i\|_2} \cdot m_i \tag{5}$$

where $O=\{o_1,...,o_N\} \in R^{N \times 3}$ is the offset vectors for $M$ points, $m=\{m_i,...,m_N\}$ is a binary mask, $c_i$ is the centroid of the instance and $Coords=\{p_i\}$ is the point coordinate set.

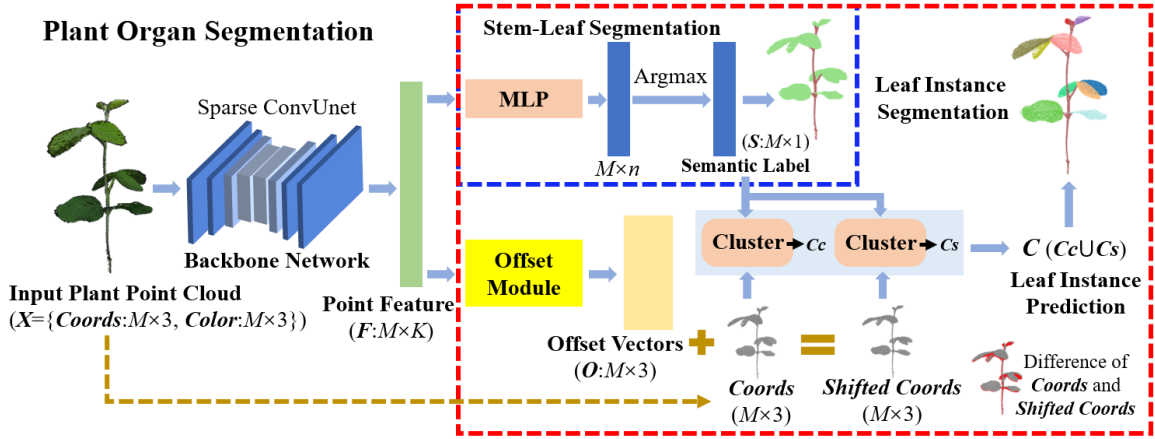

Figure 7: Illustration of the plant organ segmentation. First, a plant point cloud $X$ (Coords: $M\times3$, Color: $M\times3$) is fed into the Sparse ConvUnet to extract the point future $F$ ($M\times K$). $M$ is the number of points and K is the dimensions of point future. Second (Stem-leaf segmentation): stem leaf branch produces stem leaf semantic scores ($M\times n$) with $F$, where $n$ is the number of classes. The predicted stem leaf label $S$ ($M\times1$) for a point is the class with maximum score (Argmax). Third (Leaf instance segmentation): offset module produces offset vectors $O$ ($M\times3$). Then, a clustering method groups points into leaf clusters on original coordinate *Coords* and *Shifted Coords*, which produce $Cc$ and $Cs$ respectively. Last, we denote the union of $Cc$ and $Cs$ as the final clustering results $C$.

### 2.3.4 Plant organ segmentation inference

After obtaining the models of plant stem-leaf and leaf instance segmentation, we fed the raw plant point clouds into the trained plant stem-leaf and leaf instance networks to achieve the results of plant organ segmentation.

### 2.3.5 The network training, testing, and evaluation

For our experiments, the soybean dataset was split into training and validation sets with 120 point clouds and 25 point clouds, the tomato dataset was split into 55 point clouds and 22 point clouds. For the pre-training, we used all point clouds without any annotations to train the backbone model. After down sampling, 1024 points ($H$) were selected as the abstraction of a point cloud. The setting for feature dimension $D$ was 512. The pre-training experiments were conducted with a batch size of 2 for 10000 iterations using a NVIDIA GeForce RTX 3090 GPU. The initial learning rate was 0.1, decayed by a polynomial rule. The weakly-supervised plant segmentation experiments were conducted by weakly-annotation point clouds with a batch size of 2 for 4000 iterations using the same GPU with the same learning rate setting. For all the experiments, we used the same Sparse ConvUnet as the backbone. For both training and testing, the voxel size for Sparse ConvUnet was set to 0.2 mm for soybean and 0.5mm for tomato. The computational overhead of them on the soybean dataset were around 48 hours and 36 hours, and around 30 hours and 20 hours on the Pheno4D dataset.

In our study, the performance of plant stem-leaf segmentation accuracy analysis was evaluated using five quantitative metrics, such as Precision, Recall, F1-score, mean Intersection-over-Union (mIoU) and the IoU per class. These five metrics are defined as follows:

$$\text{Precision} = \frac{TP}{TP+FP} \quad (6)$$

$$\text{Recall} = \frac{TP}{TP+FN} \quad (7)$$

$$F1 = \frac{2 \text{ Precision} \times \text{Recall}}{\text{Precisiom} + \text{Recall}} \tag{8}$$

$$IoU = \frac{TP}{TP+FP+FN} \tag{9}$$

$$mIoU = \frac{1}{n}\sum_{i=1}^{n} \frac{TP_i}{TP_i + FP_i + FN_i} \tag{10}$$

where TP, FP and FN are the true positives, false positives and false negatives. $n$ is the number of the label categories. The mIoU is calculated by averaging the IoU over all the classes.

The performance of plant leaf instance segmentation was evaluated by Average Precision (AP). In our experiments, AP@25 and AP@50 denoted that AP scores with IoU threshold set to 25% and 50%. And AP averages the scores setting IoU threshold from 50% to 95% with the step of 5% [33].

### 2.4 Phenotypic traits extraction and evaluation

After plant organ segmentation, three plant phenotypic traits (stem diameter, leaf width, and length) were extracted (Fig. 1(C)). For the stem traits, we used the stem points of stem leaf segmentation results to calculate the stem diameter. First, we separated stem points into 4 uniform parts along z-axis. Then we fitted a straight-line segment on the part of stem points with minimum z-value using the least squares method. Last, we computed the projection distances from these stem points to this straight-line, and chose twice the median of these distances as the stem diameter [34].

Leaf length and width were calculated by each leaf instance segmentation results. First, we computed the first, second and third principal component axes of individual leaf points using the Principal Component Analysis (PCA). We found two end points along the first axis. The leaf length was obtained by the shortest path between these two end points. Second, we divided the leaf points into 5 parts along the first principal component vector. And then in each part, we found two end points along the second principal component vector and third principal component vector, respectively. We determined the longest path of the shortest path between two end points in these two groups as the leaf width [34].

The accuracy of the phenotypic traits extraction was calculated by the correlation coefficient ($R^2$) and root-mean-square error (RMSE).

$$R^2 = 1 - \frac{\sum_{l=1}^{m}(e_l - e_l')^2}{\sum_{l=1}^{m}(e_l - \bar{e}_l)^2} \tag{11}$$

$$RMSE = \sqrt{\frac{1}{m}\sum_{l=1}^{m}(e_l - e_l')^2} \tag{12}$$

where $e_l$ and $e_l'$ are the ground truth and prediction of the plant phenotypic trait; $\bar{e}_l$ is the mean of the ground truth $m$ denotes the number of the objects to be compared.

### 3. Results

#### 3.1 Plant stem leaf segmentation

The stem leaf segmentation results were assessed qualitatively and quantitatively. Fig. 8 and Fig. 9 presented the representative stem leaf segmentation results for soybean and tomato plants at different growth stages, respectively. Overall, these results showed that the proposed Eff-3DPSeg exhibited good generalization ability and accuracy for 3D plant shoot segmentation using weak supervision. From the qualitative results, it was observed that the semantic segmentation performance of all weak supervision settings was similar to the full supervision results. However, there were still some false classified points in the results. For the soybean plants, the misclassifications happened on points of the connections between a stem and a leaf and the edges of leaf as indicated in red zoomed-in boxes in Fig. 8. In the zoomed-in box *a*, some points of junctions of leaves and stem were misclassified as the category 'leaf'. In the zoomed-in box *b*, some points in the edge of green leaf were misclassified as the category 'stem'. Generally, a tomato plant has a more complex shoot structure and more leaves compared with a soybean plant in this study. Our method still showed good segmentation results on the tomato samples with both weakly supervised settings and fully supervised setting (Fig. 9). However, we had similar misclassification situations. For example, it was difficult to distinguish the exact junction between a leaf and a stem. As shown in the zoomed-in box *b* (Fig. 9), some stem points were falsely classified as leaf points. Especially in 50-points setting, some points of the main stem were even misclassified as leaf points in the zoomed-in box *e*.  For other training settings, the segmentation results of the main stem were accuracy. In boxes *c* and *d* (Fig. 9), there were some gaps in the edge of the leaf. For this situation, the points of the leaf edge were falsely classified as stem points in 200-point and 100-point settings.

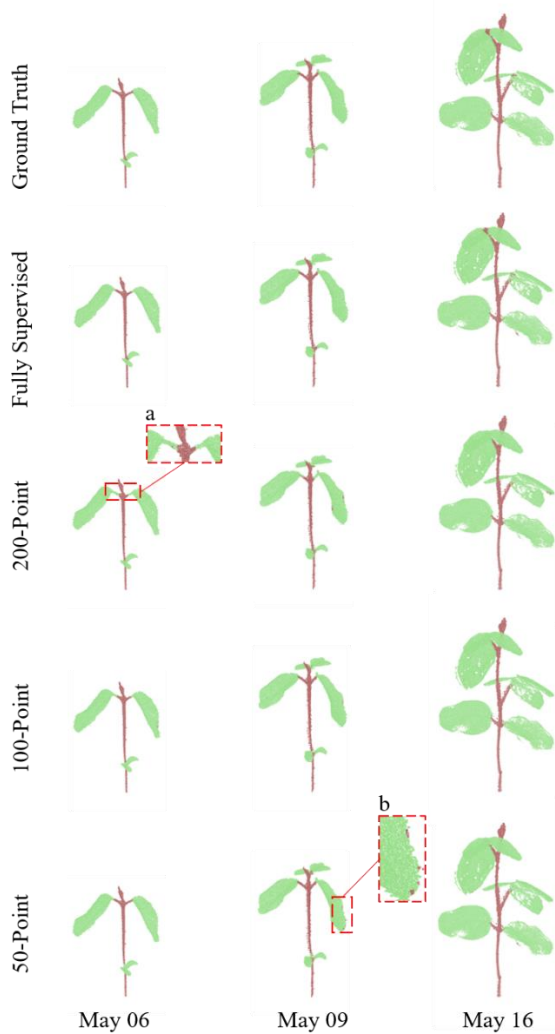

Figure 8: The qualitative visualizations for soybean stem leaf segmentation under different supervision settings. The displayed samples were selected for covering individuals that were with different grow

stages. The stem-leaf segmentation ground truth and the results of different supervision settings were shown in different rows.

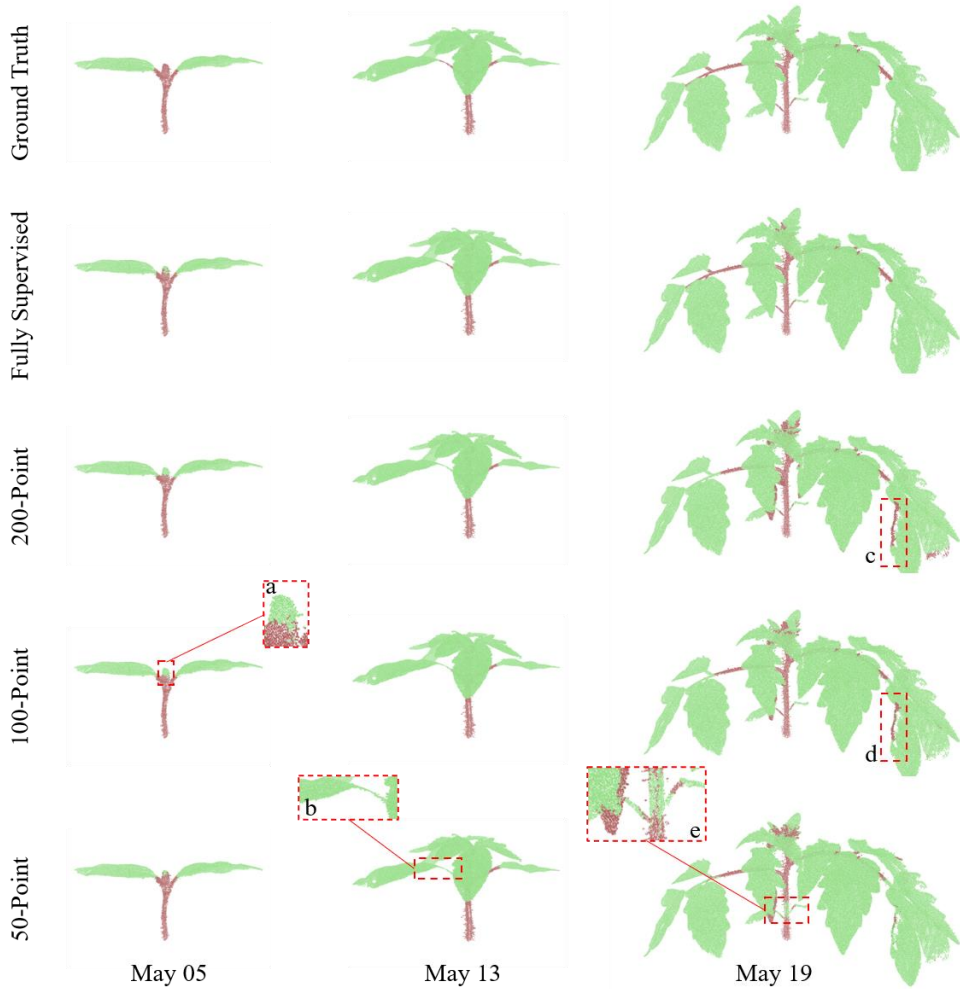

Figure 9: The qualitative visualizations for tomato stem-leaf segmentation under different supervision settings. The selected samples were with different grow stages. The stem-leaf segmentation ground truth and the results of different supervision settings were shown in different rows, respectively.

The quantitative results were summarized in Table 2 and Table 3. Baseline means the plant organ segmentation model is trained without any self-supervised representation learning pre-trained weights. In a nutshell, Eff-3DPSeg outperformed the baseline by large margins in all quantitative metrics for both soybean and tomato plants. This demonstrated that our self-supervised pre-training method obtained meaningful representation from the plant point clouds which was significantly beneficial for the weakly supervised plant organ segmentation. In the quantitative results, an interesting fact was that as the number of supervised points decreased, the margins grew larger. Meanwhile, we noticed that Eff-3DPSeg had better segmentation performance in simpler plant structures, such as the performance of leaf was better than the stem, and the soybean stem leaf segmentation results was better than that of tomato. The reasons could be that (1) the number of leaf points is larger than that of stem in the training setting and the stem spatial structure is more complex than leaf; (2) the training data for soybean plants is richer than that for tomato plants. For the network training, the amount of training data directly affects the performances of the segmentation. In Table 2, our method achieved improvements of soybean stem leaf segmentation performance in all four metrics. However, for IoU and Precision, the values of the 50-point supervision setting was larger than the 200-point setting.

In addition, we also compared with several commonly used point cloud segmentation methods under full supervision setting, including PointNet [35], PointNet++ [36], and PVCNN [37] in Table 2 and 3. It was observed that our method achieved the best performance among these methods. The means of Precision, Recall, F1-score and IoU were all about 1% positive margins than the second-best method for both soybean and tomato datasets.

Table 2. Weakly supervised plant stem-leaf segmentation results of soybean plants

| Supervison | Method | Precision(%) | | | Recall(%) | | | F1-score(%) | | | IoU(%) | | |
|---|---|---|---|---|---|---|---|---|---|---|---|---|---|
| | | stem | leaf | mean | stem | leaf | mean | stem | leaf | mean | stem | leaf | mean |
| 50 | Baseline | 81.1 | 99.5 | 90.3 | 96.7 | 96.9 | 96.8 | 88.2 | 98.2 | 93.2 | 78.9 | 96.5 | 87.7 |
| | Eff-3DPSeg | 95.2 | 98.9 | 97.0 | 93.6 | 99.2 | 96.4 | 94.4 | 99.1 | 96.7 | 89.3 | 98.1 | 93.7 |
| 100 | Baseline | 90.4 | 99.3 | 94.9 | 95.6 | 98.4 | 97.0 | 92.9 | 98.9 | 95.9 | 86.8 | 97.7 | 92.3 |
| | Eff-3DPSeg | 93.8 | 99.2 | 96.5 | 95.4 | 99.0 | 97.2 | 94.6 | 99.1 | 96.8 | 89.7 | 98.2 | 94.0 |
| 200 | Baseline | 89.6 | 99.1 | 94.4 | 94.3 | 98.3 | 96.3 | 91.9 | 98.7 | 95.3 | 85.0 | 97.4 | 91.2 |
| | Eff-3DPSeg | 91.0 | 99.2 | 95.1 | 94.7 | 98.5 | 96.6 | 92.8 | 98.8 | 95.8 | 86.7 | 97.7 | 92.2 |
| Full | PointNet | 37.5 | 89.0 | 63.2 | 33.9 | 90.4 | 62.2 | 35.6 | 89.7 | 62.6 | 21.7 | 81.3 | 51.5 |
| | PointNet++ | 90.2 | 99.4 | 94.8 | 95.6 | 98.5 | 97.1 | 92.8 | 98.9 | 95.9 | 86.6 | 97.9 | 92.3 |
| | PVCNN | 91.4 | 99.5 | 95.5 | 96.8 | 98.7 | 97.7 | 94.0 | 99.1 | 96.6 | 88.7 | 98.3 | 93.5 |
| | Eff-3DPSeg | 98.5 | 99.5 | 99.0 | 96.8 | 99.8 | 98.3 | 97.7 | 99.6 | 98.6 | 95.4 | 99.2 | 97.3 |

Table 3. Weakly supervised plant stem-leaf segmentation results of tomato plants (Pheno4D)

| Supervison | Method | Precision(%) | | | Recall(%) | | | F1-score(%) | | | IoU(%) | | |
|---|---|---|---|---|---|---|---|---|---|---|---|---|---|
| | | stem | leaf | mean | stem | leaf | mean | stem | leaf | mean | stem | leaf | mean |
| 50 | Baseline | 76.6 | 96.3 | 86.4 | 75.3 | 96.5 | 85.9 | 75.9 | 96.4 | 86.2 | 61.2 | 93.1 | 77.1 |
| | Eff-3DPSeg | 86.6 | 97.1 | 91.8 | 81.4 | 98.0 | 89.7 | 83.9 | 97.5 | 90.7 | 72.3 | 95.2 | 83.7 |
| 100 | Baseline | 66.4 | 95.8 | 81.1 | 69.9 | 95.1 | 82.5 | 68.1 | 95.4 | 81.7 | 51.6 | 91.2 | 71.4 |
| | Eff-3DPSeg | 86.8 | 98.1 | 92.4 | 86.9 | 98.0 | 92.5 | 86.9 | 98.1 | 92.5 | 76.8 | 96.2 | 86.5 |
| 200 | Baseline | 87.2 | 98.3 | 92.8 | 88.5 | 98.1 | 93.3 | 87.8 | 98.2 | 93.0 | 78.3 | 96.5 | 87.4 |
| | Eff-3DPSeg | 91.5 | 98.0 | 94.7 | 87.0 | 98.7 | 92.9 | 89.2 | 98.4 | 93.8 | 80.5 | 96.8 | 88.6 |
| Full | PointNet | 54.2 | 91.1 | 72.6 | 53.4 | 91.3 | 72.4 | 53.8 | 91.2 | 72.5 | 36.8 | 83.8 | 60.3 |
| | PointNet++ | 85.9 | 99.2 | 92.5 | 95.2 | 97.4 | 96.3 | 90.3 | 98.3 | 94.3 | 82.3 | 96.6 | 89.4 |
| | PVCNN | 84.4 | 98.5 | 91.4 | 91.1 | 97.1 | 94.1 | 87.6 | 97.8 | 92.7 | 78.0 | 95.6 | 86.8 |
| | Eff-3DPSeg | 94.7 | 99.4 | 97.1 | 95.9 | 99.2 | 97.5 | 95.3 | 99.3 | 97.3 | 91.0 | 98.6 | 94.8 |

## 3.2 Plant leaf instance segmentation

The leaf instance segmentation performance for soybean and tomato plants under different growth stages using different number of supervision points were also assessed qualitatively and quantitatively. Fig. 10 and Fig. 11 presented the representative results of plant leaf instance segmentation. Like the stem leaf segmentation, it was observed that the network training with weak supervision points achieved nearly the same performance as that of fully supervised setting. However, because of dense leaves and limited training samples, misclassifications also happened on points of the edges of leaf and connections between a stem and a leaf. For example, the zoomed-in box *a* in Fig. 10 and Fig. 11, some points of the edge of leaves were falsely classified. Due to the gaps in soybean leaves, the individual leaf was incorrectly clustered several parts as shown in boxes *a*, *b*, and *c* (Fig. 10). Even in

Page 13 of 21

box *b*, some points of the stem were clustered as a part of a leaf. For tomato (Fig. 11), three closely leaves were segmented as the same leaf under the100-points and the 50-points settings in boxes *d* and *f*. For box *e*, the edge points of the leaf were misclassified under the 100-point setting and one leaf was clustered into several parts under the 50-point supervision in box *g*. While these were correct in the 200-point setting, except for points of the top stem which were falsely clustered as a leaf in box *b*. This demonstrated that the performance of the leaf instance segmentation improves with increasing the number of supervision points.

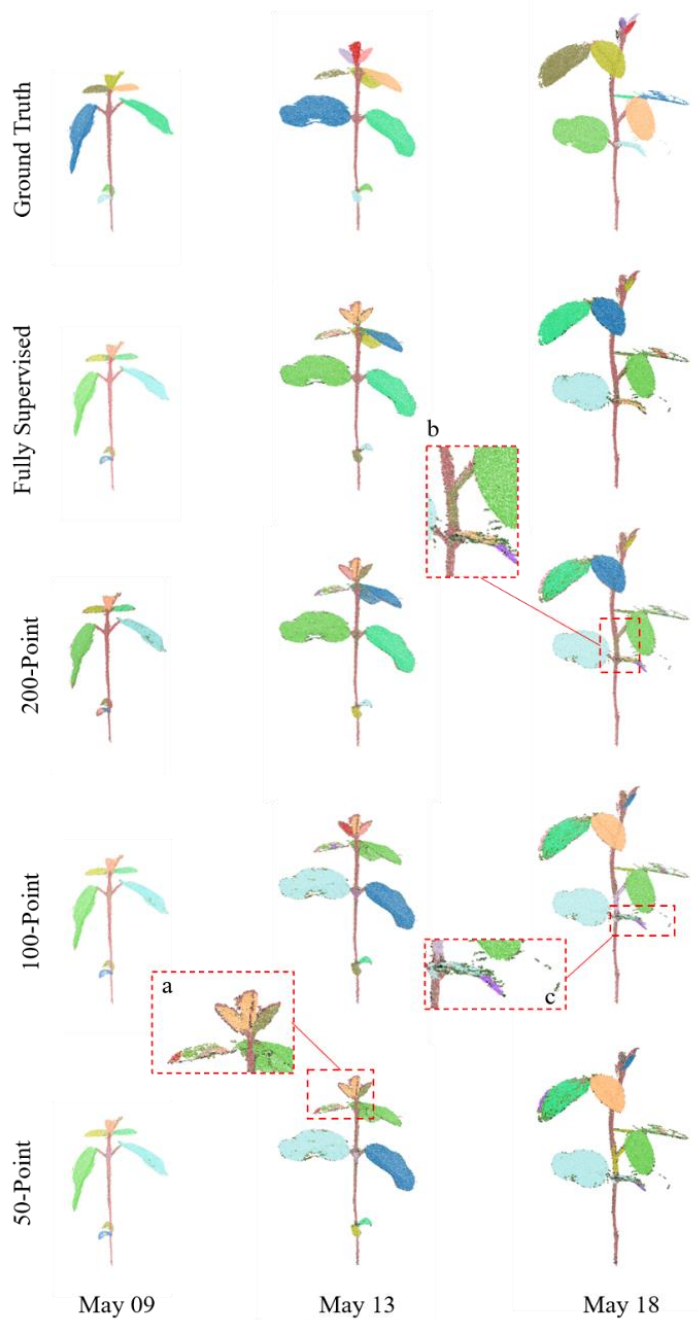

Figure 10: The qualitative visualizations for weakly-supervised soybean leaf instance segmentation. The samples were with different grow stages. The 5 rows showed the leaf instance segmentation ground truth and results of different supervision settings, respectively.

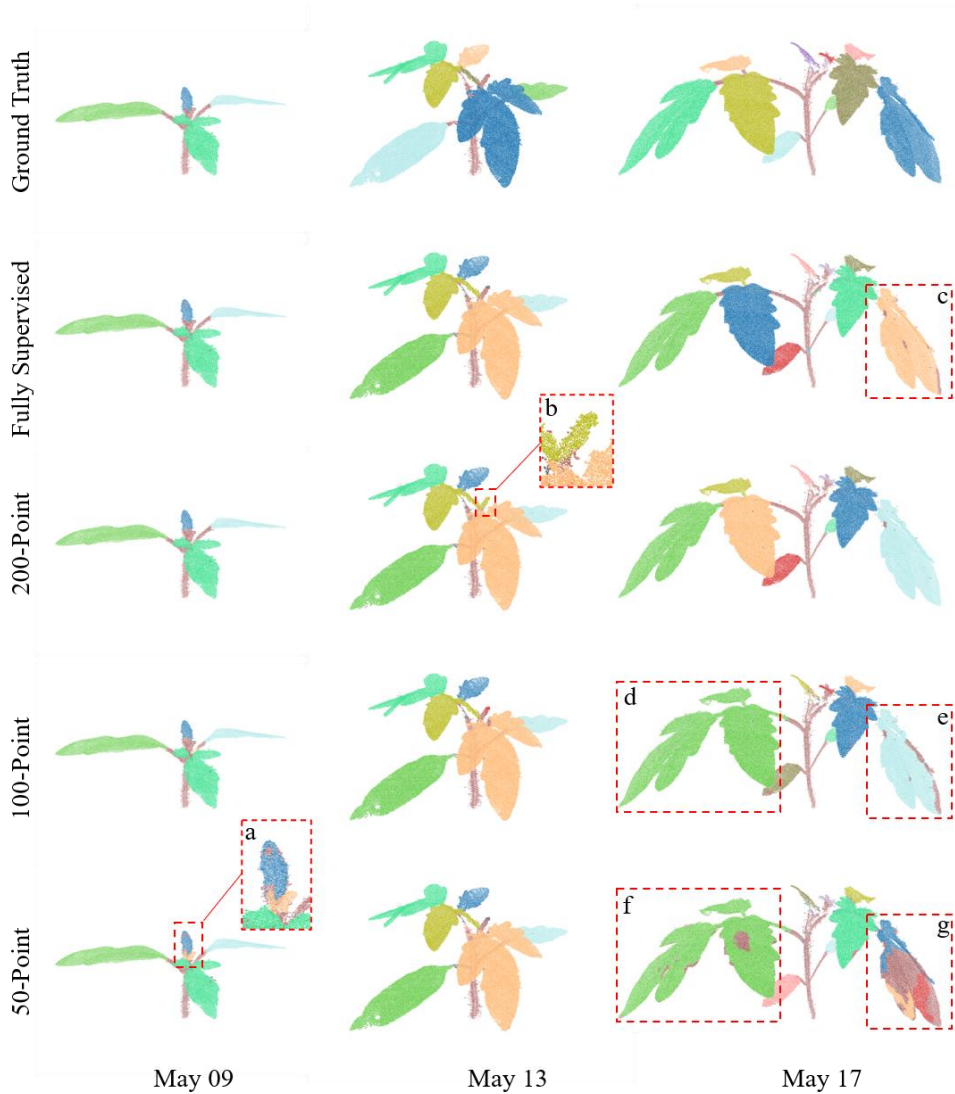

Figure 11: The qualitative visualizations for weakly-supervised tomato leaf instance segmentation. The selected tomato samples were with different grow stages. And tomato leaf instance segmentation ground truth and results of different supervision settings were shown in different rows.

Table 4 summarized the quantitative results of the soybean and tomato leaf-instance segmentation. Compared with the baseline, Eff-3DPSeg achieved large positive margins for most of the supervision settings. For the soybean leaf instance segmentation, the performance of the 100-points setting achieved the best results, about 26.6 AP@50. For the tomato, the 200-points setting obtained the best performance for the leaf instance segmentation, about 53.0 AP. Similar to the stem leaf segmentation, with the number of supervision points decreased, the margins of performance with Eff-3DPSeg and baseline increased. However, we noticed that the weakly supervised results of the tomato leaf instance segmentation in the 200-points setting declined in AP@50 and AP@25, increased in AP compared with the full supervision. The reason may be that the amount of tomato training data is not enough, only 55 point clouds, leading to the trained model without better statistical property. In all supervision settings, the segmentation performance of tomato was general better than that of soybean.

Meanwhile, we also compared our method with other point cloud segmentation methods with full supervision, including PointNet and PVCNN. Our method achieved the best performance among these methods, with large margins about over 30 AP@50.

Table 4: Weakly supervised plant leaf instance segmentation results

| Supervision | Method | Soybean | | | Tomato | | |
|---|---|---|---|---|---|---|---|
| | | AP | AP@50 | AP@25 | AP | AP@50 | AP@25 |
| 50 | Baseline | 7.3 | 14.6 | 25.9 | 26.1 | 45.1 | 61.6 |
| | Eff-3DPSeg | 13.8 | 23.9 | 33.6 | 43.0 | 60.3 | 69.7 |
| 100 | Baseline | 13.1 | 22.1 | 29.4 | 30.6 | 45.9 | 61.6 |
| | Eff-3DPSeg | 14.1 | 26.6 | 38.4 | 43.8 | 62.7 | 74.2 |
| 200 | Baseline | 17.6 | 13.2 | 19.1 | 45.1 | 64.5 | 74.7 |
| | Eff-3DPSeg | 13.8 | 24.2 | 33.4 | 53.0 | 62.8 | 70.3 |
| Full | PointNet | – | 19.1 | – | – | 27.3 | – |
| | PVCNN | – | 31.0 | – | – | 35.3 | – |
| | Eff-3DPSeg | 53.6 | 64.4 | 69.3 | 57.2 | 67.6 | 73.2 |

## 3.3 Evaluation of extracted traits

The accuracy of extracted plant phenotypic traits was evaluated with the correlation coefficient $R^2$ and root-mean-square error (RMSE) (Table 5 and 6). We selected 9 soybean point clouds and 11 tomato point clouds to compare with the manual measurements and the extracted traits based on the segmentation results. Generally, our method achieved great performance for extracting plant organ-level phenotypic traits based on the proposed 3D deep learning plant organ segmentation method. However, there were opposite trends for the soybean and tomato datasets. For the tomato plants, the best performance was achieved with the 200-point setting; in contrast, the best performance for the soybean plants was achieved with the 50-point setting. That is because the accuracy of traits extraction was dependent on the performance of the plant organ segmentation. These trends were the same as the results of the plant organ segmentation described in Sections 3.1 and 3.2.

For the stem level, the stem diameter was extracted by the results of stem leaf segmentation. Depending on the high performance of the tomato and soybean stem leaf segmentation, high performance was achieved for both types of the plants in terms of the two evaluation metrics. However, the stem diameter $R^2$ and RMSE for the tomato plants were better than those for the soybean plants. That is because the ghost noisy points on the soybean point clouds affected the performance of the trait extraction. As shown in Fig. 12, the zoomed-in region 3 was the part of soybean stem, in which many ghost points were on the left of the stem. In contrast, the resolution of tomato stem was very high, the details of the stem were displayed clearly in the zoomed-in area 1 (Fig. 12). The results of the leaf width and length were depended on the leaf instance segmentation. The performance of extracted leaf phenotypic traits for the tomato plants was better than that for the soybean plants. On the other hand, there were some gaps in the soybean leaves (zoomed-in region 2 in Fig. 12), which affected the accuracy of the extracted leaf length and width.

Table 5: R2 of plant phenotypic traits extraction

| R2 | stem diameter | | leaf width | | leaf length | |
|---|---|---|---|---|---|---|
| | soybean | tomato | soybean | tomato | soybean | tomato |
| 200 points | 0.94 | 0.99 | 0.89 | 0.94 | 0.86 | 0.99 |
| 100 points | 0.95 | 0.98 | 0.89 | 0.94 | 0.90 | 0.98 |
| 50 points | 0.97 | 0.96 | 0.92 | 0.91 | 0.92 | 0.93 |

Table 6: RMSE of plant phenotypic traits extraction

| RMSE | stem diameter | | leaf width | | leaf length | |
|---|---|---|---|---|---|---|
| | soybean | tomato | soybean | tomato | soybean | tomato |
| 200 points | 0.16 | 0.01 | 5.50 | 2.34 | 4.89 | 0.99 |
| 100 points | 0.14 | 0.07 | 5.43 | 2.29 | 4.08 | 1.35 |
| 50 points | 0.11 | 0.12 | 4.58 | 2.60 | 3.59 | 2.47 |

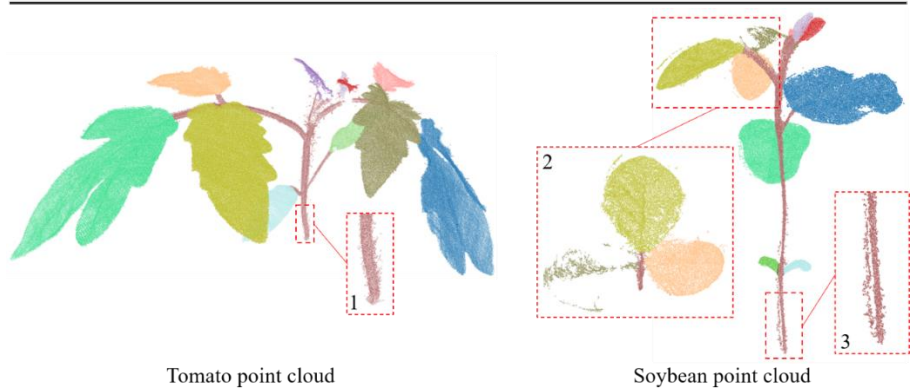

Figure 12 Comparison between the tomato point cloud (Pheno4D dataset) and the soybean point cloud (our dataset).

## 4. Discussion

### 4.1 Plant organ segmentation

Overall, the proposed weakly-supervised framework Eff-3DPSeg achieved promising performance of plant organ segmentation for soybean and tomato plants under different growth stages, the performance was close to that of the fully supervised setting (Table 2~6). It was observed that the misclassification mainly happened on the top sprouts of plants, the junction of the stems and leaves, and the edge of the leaves. This could be caused by the following reasons. First, in our experiments, we randomly chose 50, 100, and 200 points from the point clouds for weak annotation. The selected points may not be the best subset for the weakly supervised segmentation tasks. Second, the training dataset is not big enough [31]. More data under different treatments will be included in the future. The aforementioned reasons would result in the nonuniform distribution of labeled points which cannot cover representative points of the plants.

The plant leaf instance segmentation network could obtain the stem-leaf and leaf instance segmentation simultaneously (Fig. 7). However, the stem leaf segmentation performance obtained from the leaf instance segmentation network (Table 7) was not as good as that obtained from the network dedicated to the stem-leaf segmentation task (Table 2 and Table 3). Because the leaf instance segmentation network made a trade-off of the stem leaf segmentation and leaf instance segmentation, the weight of semantic segmentation part of loss function was only 1/3. In addition, the clustering part of instance segmentation depends on the outputs of two branches (Fig. 7). If the performance of stem leaf segmentation is worse, it will affect the final results of the leaf instance segmentation. Hence, we will optimize the leaf instance segmentation framework. Firstly, we will fuse the outputs of two branches to carry out effective information interaction between the semantic and instance feature map. Then we will improve the loss function of stem leaf and leaf instance segmentation to implement high performance of semantic and instance segmentation results, simultaneously.

Table 7: The stem-leaf segmentation results of Eff-3DPSeg leaf instance network (IoU: %)

| NO. of annotated points | Soybean | | | Tomato | | |
| :---: | :---: | :---: | :---: | :---: | :---: | :---: |
| | stem | leaf | mIoU | stem | leaf | mIoU |
| 50 | 72.5 | 94.6 | 83.6 | 34.1 | 83.2 | 58.6 |
| 100 | 83.0 | 96.7 | 89.8 | 47.8 | 86.6 | 67.2 |
| 200 | 79.4 | 96.0 | 87.7 | 50.9 | 87.5 | 69.2 |

For another similar 3D plant organ segmentation work PlantNet [25], it developed a fully-supervised deep learning network for plant semantic and instance segmentation. The network required that the input point cloud has a fixed number of points (4096). In addition, the input point cloud only contains the XYZ 3D coordinates. However, our method is flexible to the dimension of the input point cloud. For example, our soybean dataset includes both XYZ 3D coordinates and RGB color information, and the Pheno4D dataset only contains 3D coordinate information (i.e., tomato). The ability allowing higher dimension input could be beneficial for improving the performance of plant organ segmentation by fusing new features. For example, we may fuse thermal or multispectral information to the point clouds in the future, and the new features would be useful for the segmentation tasks. Moreover, our method does not require the input point cloud with a fixed point number. Our network could conduct the huge computation for point clouds with more than 100,000 points, which is particularly useful for processing plants with large size shoots such as maize plants and trees. Additionally, our weakly supervised plant organ segmentation framework, Eff-3DPSeg, only needs to label around 0.5% of points, which can dramatically save the annotation time.

### 4.2 Plant phenotypic traits extraction

Based on the results of plant organ segmentation, we achieved good performance for plant organ phenotypic traits extraction for both datasets. However, it was observed that the results for the tomato dataset were slightly better than those for our soybean dataset. In addition to the performance difference of the plant organ segmentation for the two datasets, another reason is about the quality of plant point clouds. Our soybean dataset was reconstructed by a low-cost photogrammetry system (MVSP2), the total cost was around $1,500. In contrast, the Pheno4D (tomato) dataset [28] was built by a light section scanner coupled to an articulated measuring arm, which could generate high-resolution plant point clouds, but costed more than $ 20,000. We will add one more RGB camera for our imaging system to improve the point color quality.

### 4.3 Limitations and future works

Our framework, Eff-3DPSeg, achieved the promising performance for 3D plant shoot segmentation. Nevertheless, there are still some limitations in our method. First, some of the soybean data captured using the MVSP2 have gaps (missing points) on leaves (such as the point clouds captured on May 18 and 25 in Fig. 2), which affects the leaf instance segmentation. Additionally, the ghost noisy points (the point clouds captured on May 16 in Fig. 2), also could affect the trait measurements. In the future, we will develop point cloud pre-processing methods for gaps filling and denoising. Second, our deep learning network needed to train the stem-leaf segmentation and leaf instance segmentation separately. Although we can obtain the results of stem leaf segmentation in the leaf instance segmentation processing, the performance of stem leaf segmentation was worse than the performance of the model that was trained directly by the stem-leaf segmentation network.

We will improve the framework such as optimizing the loss function for the stem-leaf segmentation part to implement the stem-leaf and leaf instance segmentation simultaneously and increase the performance. Third, we provided two categories (stem and leaf) in this study. In the future, we can classify a plant intensively into more categories such as leaf, main stalk, branch, petiole, and growing point based on desired plant breeding objectives. Also, we will produce more point clouds with various plant species to further test our method, improving its generality and diversity.

5. **Conclusion**

In this study, we proposed a novel annotation-efficient deep learning framework, Eff-3DPSeg, for 3D plant shoot segmentation. Firstly, we developed a low-cost multi-view imaging data acquisition platform (MVSP2) and a point cloud annotation tool (MPA) to build a spatio-temporal point cloud dataset for soybean plants. Then, three different annotation settings (50, 100, 200 annotated points) for the soybean dataset and a public dataset Pheno4D were used to train and test the proposed network Eff-3DPSeg. Overall, our method achieved similar plant organ segmentation performance in 3D compared with the fully supervised setting, and then three organ level phenotypic traits were well extracted. In addition, on the one hand, our method can dramatically save point cloud annotation time; on the other hand, the point cloud reconstruction can be achieved using a low-cost multi-view imaging platform. We believe that this work will contribute to the efficiency of high throughput plant phenotyping and the development of smart agriculture.

**Acknowledgement:**

This work is supported by the Natural Sciences and Engineering Research Council of Canada (NSERC) Discovery Grants Program (Grant no. G256643), Fonds de Recherche du Québec Nature et technologies (FRQNT) Programme de recherche en partenariat - Agriculture durable (Grant No. G259806 FRQ-NT 322853 X-Coded 259432) and FRQNT Emerging project (2022-AD-309895).

**Author contributions:**

S.S, M. L, and L.L conceived the idea and designed the experiments. L.L developed and tested the method, conducted experiments, analyzed the results, and wrote the original draft. X. J, Y. Y and E. R. A. S contributed to growing plants, image data acquisition, point cloud reconstruction and annotation. S.S, M.L, and V.H.V conducted the supervision and performed revisions of the manuscript. All authors contributed to editing, reviewing, and refining the manuscript.

**Declaration of Competing Interest:**

The authors declare that there is no conflict of interest regarding the publication of this article.